%% file: main.tex
\documentclass[journal]{new-aiaa} 
\usepackage[utf8]{inputenc}
\usepackage{comment}
\usepackage{graphicx}
\usepackage{amsmath}
\usepackage[version=4]{mhchem}
\usepackage{subcaption}
\usepackage{float}
\usepackage{pgfplots}
\usepackage{tikz}
\usetikzlibrary{shapes,arrows}
\usepackage{booktabs}
\usepackage{dirtytalk}

\usepackage{soul}

\usepackage{siunitx}
\usepackage{longtable,tabularx}
\usepackage{caption}
\title{{Training Observable Control Policies to Expose Agent State Through Actions}
\footnote[3]{Estimating System State from the Actions of a Reinforcement Learning Agent, AIAA 2023-2657, Session: Autonomy V, Published Online:19 Jan 2023https://doi.org/10.2514/6.2023-2657, AIAA Scitech 2023 Forum, 23-27 January 2023, National Harbor, MD \& Online}
\footnote[4]{Enabling Inter-Vehicle Coordination Through Observable Control Policies, AIAA 2024-0989, Session: Autonomy, Published Online: 4 Jan 2024https://doi.org/10.2514/6.2024-0989, AIAA Scitech 2024 Forum, 8-12 January 2024, Orlando, FL}}

\author{Andres Enriquez Fernandez\footnote{Graduate Research Assistant, Department of Aerospace and Mechanical Engineering, 500 W University, El Paso, TX 79968, aenriquezf@miners.utep.edu, AIAA Member.}}
\author{John J. Bird\footnote{Assistant Professor, Department of Aerospace and Mechanical Engineering, 500 W University, El Paso, TX 79968, AIAA Member.}}
\affil{The University of Texas at El Paso, El Paso, TX, 79968}

\begin{document}

\maketitle




\input{abstract_v2}

\input{introduction_v2}

\input{approach_v2}

\input{experiments_v2}

\input{results_v2_v2}

\input{discussion_v2}

\input{conclusion_v2}

\section*{Funding Sources}
Portions of this work were funded by the Air Force Office of Scientific Research under award FA9550-24-1-0176. 

\bibliography{andres.bib, publications-scitech_2023_learning, bird2.bib}

\end{document}

%% file: abstract_v2.tex
\begin{abstract}

Physical or operational constraints often impose communications limitations on autonomous agents. Such limitations complicate monitoring or multiagent coordination. Even when strong communications are absent, some information may still be available. The remainder of the relevant agent state may be reconstructed via estimation. The actions taken by an agent are a potential source of information -- as the agent interacts with the environment, these actions may be observed even in the absence of explicit communication. We investigate using actions to estimate the state of an agent, using reinforcement learning to develop policies which make the estimation problem more tractable. Policy observability is encouraged through the training reward and is analyzed using simulation of the trained agent. In an aircraft tracking problem a policy with enhanced observability is found that has minimal impact on nominal task performance.

\end{abstract}

%% file: introduction_v2.tex
\section{Introduction}

Effective collaboration between cooperating agents (including autonomous to autonomous agents and human to autonomous agent coordination) typically depends on a mutual understanding of at least a subset of the states of each agent. 
This is often achieved via direct communication that enables complex behavior, coordination, environmental information sharing, and sub-task learning \cite{panait2005cooperative,dorri2018multi}. 
Reliability and availability of communications is not assured however.
Communications limitations may appear at the channel level as delays, drops, and bandwidth limitations.
Limited or absent direct communication can also occur because of physical or operational considerations.
The environment may inhibit communication (e.g. underwater), the operation may be compromised by excessive communication, or the agents may not be equipped with common hardware or language for direct communication.

Casting the problem as a partially observable Markov Decision Process (POMDP) offers one approach to resolve uncertainty introduced when states are not directly communicated. 
Under this formulation, the state vector of every agent includes the agent's belief of the state of the other agents \cite{roth2006communicate}, and the agents are rewarded for their collective actions \cite{liu2016learning}. 
This couples the decision-making model to the collective actions of all the agents. 
One implication of this structure is that changes to the task of even a single agent requires retraining all the agents, limiting practicality in a federated system. 
Another practical constraint is that the POMDP approach scales poorly as the dimension of joint action, state, and observation spaces grow exponentially with the number of agents \cite{amato2015scalable}. 

When coordination is approached as a POMDP, an observer is implicitly developed in the agent through the training process \cite{murphy2000survey}. An alternative which ``decouples'' the agents is to adopt the observer-controller structure common in control systems, explicitly formulating an estimator rather than training it implicitly through the expanded agent state. Here, the estimator assimilates a subset of the available information to infer the needed states of other agents, and enable coordination \cite{8814643}. One option for the observations are the control actions taken by agents. These observations have the advantage that an observation model is already defined (the control policy), and that in some cases they may be visible through direct observation without explicit communication; in effect they may be ``broadcast'' for free.

Informal examples of such implicit information sharing through the observation of actions occur in a number of contexts. For example, a ground vehicle rapidly braking or changing lanes can indicate an obstacle in the road, an aircraft observed to make a course change in the presence of nearby traffic has likely detected it visually, and an aircraft which deploys its undercarriage is likely near an airport. Enabling this implicit communication, and increasing its efficiency can enable coordination between autonomous agents in communications limited scenarios. It may also enhance the comfort and safety of mixed human-autonomous operations as human agents are more likely to understand what the agent is doing.

To improve observability of an agent's control policy, we formulate an estimator which uses as observations the actions chosen by a control policy. We train the policy to reward the estimator's performance while the agent performs a target tracking task. The observable policy is compared to one trained only to perform the tracking task, comparing task-only performance, estimate quality, and a measure of observability. 

While one objective of this work is to enable coordination in communication-limited scenarios, we focus here on estimating the state of a single agent given only a limited stream of observations from that agent. If the state can be determined reliably, then collaborative control policies can be defined using the estimated states \cite{yang2008multi,smith2007closed}. The result is a scenario in which agents can be trained and updated separately, provided their control policies can be shared with other agents prior to a task.

Contributions of this work include: 1) we implement an estimator for an agent's state that uses only its control policy output as observations, 2) we improve observability of control policies using reinforcement learning, 3) we demonstrate through singular value decomposition analysis of the observability matrix and through Monte Carlo simulation that the resulting policies improve estimator performance with minimal impact on the nominal task performance.

\begin{figure}
    \centering
    \includegraphics{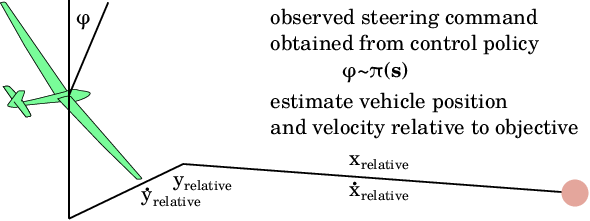}
    \caption{The example problem used to explore training observable control policies. The estimator must determine the agent's state (position and velocity relative to the target) with observation only of the steering command, $\phi$. }
    \label{fig:problem}
\end{figure}

Figure \ref{fig:problem} illustrates our test problem -- a control policy must keep a fixed-wing aircraft near a target point by controlling the aircraft bank angle.
A second agent tries to determine the first agent's position $(x_{relative},y_{relative})$ and velocity $(\dot{x}_{relative},\dot{y}_{relative})$ relative to the target point using as an observation only the steering command selected by the first agent. 

%% file: approach_v2.tex
\section{Agent Definition}

Decision-making agents can broadly be described with a control policy -- a mapping that associates the agent's state (which can include both physical state and the internal information state of the agent) with a control action to be executed. The agent has the objective of steering the system from some state $\mathbf{s}_k$, at step $k$, to a future state or set of states $\left\{ \mathbf{s}_{objective} \right\}$. This model can describe a wide variety of systems, from classical control systems to human operators \cite{kelly2019hg}. Control policies can be constructed via direct mathematical analysis or data-driven approaches, and can accommodate both deterministic and probabilistic agents and control approaches. We specifically focus on probabilistic agents, which associate states with distributions over actions:
\begin{equation*}
    \pi: \mathbf{s}_k \rightarrow p(\mathbf{a})~\forall~\mathbf{a} \in \mathcal{A}(\mathbf{s}_k)
\end{equation*}
where $\pi$ is the agent's policy function, $\mathbf{a}$ is an action, $p(\mathbf{a})$ is the probability density function associated with the actions, and $\mathcal{A}(\mathbf{s}_k)$ is the set of all actions which are permissible in state $\mathbf{s}_k$. The agent samples an action $\tilde{\mathbf{a}}\sim p(\mathbf{a})$ which is executed during each decision interval. Thus, for continuous states and actions the control is a stochastic function that maps $\mathbb{R}^n\rightarrow\mathbb{R}^m$ with $n$ the dimension of the state space and $m$ the dimension of the action space.

We represent the policy function with a neural network that we train using reinforcement learning and a simulation environment. The input of the neural network is the state of the system, and the outputs are parameters of a pseudocontrol distribution which is sampled to obtain control variables. The pseudocontrol to control mapping is given by:
\begin{equation}
    \mathbf{a}_i = \tanh(\mathbf{u}_i) \mathbf{a}_{i, saturation} 
    \label{eq:pseudo_mapping}
\end{equation}
where $\mathbf{u}_i$ are elements of the sampled pseudocontrol vector which are multiplied by the control limits, $\mathbf{a}_{i, saturation}$. The sampled control actions influence the evolution of the agent's dynamics both during training and evaluation. The sampled actions are also made available to an estimation system that aims to estimate the state of the agent.

\section{Estimator and Reward Embedding}

We formulate an estimator using the Unscented Kalman filter (UKF) \cite{simon2006optimal}, which is well suited to handle the nonlinearities in the system dynamics and control policies. The filter propagates the state estimate forward in time using known dynamics and the expected control action while assimilating the agent's pseudocontrol actions $\tilde{\mathbf{u}}$ as observations. This pseudocontrol is sampled from the agent's policy output distribution given its current state $\mathbf{s}_k$. The control policy can thus be interpreted as the filter's measurement function, and the action distribution at each state characterizes the observation noise. The measurement update innovation is thus the difference between the actual and expected control action at the expected state:
\begin{align*}
     \mathbf{z} &= (\tilde{\mathbf{u}} - \hat{\mathbf{u}}) \nonumber \\
                &= (\tilde{\mathbf{u}} - \mathbb{E}(\mathbf{u})) \nonumber \\
                &= (\tilde{\mathbf{u}} - \mathbb{E}(\pi(\hat{\mathbf{s}}_k))
\end{align*}
where $\mathbf{u}$ is a random variable distributed according to the policy function, $\tilde{\mathbf{u}}$ is the observed realization of the action distribution, and $\hat{\mathbf{u}}$ is the expectation of the action at the current state estimate.
We assume that the estimator has a priori knowledge of the system dynamics and the control policy. However, at run time the only information available to the estimator is the agent's pseudocontrol actions.
Prior tests with this filter formulation \cite{doi:10.2514/6.2023-2657} show that the estimator does not perform well for all policies. To resolve this challenge, we embed the estimator into the training environment, simulating the evolution of the estimated state as well as the actual state of the system. The estimator performance is then available as a reward signal to augment the task-only reward:
\begin{equation}
     r = r_{task-only} + \mathcal{R}(\hat{x}, x)
\label{eq:explicit_reward}
\end{equation}
where $r$ is the reward for a given task and $\mathcal{R}(\hat{x}, x)$ is a reward function that promotes actions which drive the estimated state to the true state. The policy now seeks to maximize task performance as well as select actions which improve estimator performance. Note that the sum allows equivalent (or greater) total reward to be achieved even if a policy results in a reduction in the task-only reward. Thus, improving the estimator performance may come at a cost to the performance on the objective task.

\subsection{Observability}

The performance of an estimator can be diagnosed through the distribution of the error between the expected and true state. While this provides a very direct measure performance, the true state must be available.
Additionally, determining why the source of error may be difficult as it can arise from sensor quality, system error, or because the available observations provide poor observability of the state.
The question of observability is important because it places mathematical constraints on the performance of any estimator regardless of structure or algorithm.
While observability for nonlinear systems is difficult to compute analytically, we can linearize a discrete nonlinear system:
\begin{align*}
    \dot{x}_k &= f_{k-1}(x_{k-1}) \\
    y_k &= h(x_k)
\end{align*}
using Taylor series approximation about state $x_{k-1}$ for the state equation and about $x_{k}$ for the measurement model equation:
\begin{align*}
    x_k &\approx f_{k-1}(x_{k-1}) + A_{k-1} (x - x_{k-1}) \\
    y_k & \approx h(x_{k}) + C (x - x_{k})
\end{align*}
where $A_{k-1} = \dfrac{\partial f_{k-1}}{\partial x}\biggr \rvert_{x_{k-1}}$ and $C = \dfrac{\partial h_k}{\partial x}\biggr \rvert_{x_{k}}$. At each state, an approximate observability matrix can be computed.
\begin{equation}
\mathcal{O}_r=\begin{bmatrix}
    C_r^\intercal  & (C_r A_r)^\intercal & (C_r A_r^2)^\intercal & \dots & (C_r A_r^{n-1})^\intercal 
\end{bmatrix}^\intercal
\label{eq:LOM}
\end{equation}
where $C_r$ is a linear approximation to the observation model, computed by taking the gradient of the nonlinear observation model with respect to the state vector at time $r$, $A$ is analogously, a linearization of the nonlinear system dynamics at time $r$, and $n$ is the dimension of the state vector. Observability at time $r$ can be assessed through the singular values of $\mathcal{O}_r$.
In the linear case, this observability assessment is definitive -- if the smallest singular values of $\mathcal{O}_r$ are close to zero, then some directions in the state-space are not observable.

In the nonlinear case, the observable subspace can vary depending on the state; if the observable subspace evolves such that it covers the full state-space before information about the system degrades, the system can be overall observable in a dynamic sense \cite{Powel2020}. The observability of a sequence of states $\{\mathbf{s}_1,\mathbf{s}_2,\dots,\mathbf{s}_r\}$ can be approximately evaluated through the stripped observability matrix (SOM) \cite{li2012observability}. 
\begin{equation}
\mathcal{O}_s(r)=\begin{bmatrix}
    \mathcal{O}_1^\intercal & \mathcal{O}_2^\intercal & \dots & \mathcal{O}_r^\intercal
\end{bmatrix}^\intercal
\label{eq:SOM}
\end{equation}
where $\mathcal{O}_r$ is the approximate observability matrix, obtained by linearizing the system at each time step as described in Equation \ref{eq:LOM}. Observability can be approximately assessed for the state trajectory through the singular values of $\mathcal{O}_s$.
This measure of observability can be used to explore the impact that rewarding estimator performance has on the control policy.

%% file: experiments_v2.tex
\section{Simulation Experiments}

We aim to produce a control policy (observation model) that allows the agent state to be reconstructed from observations of the control actions. To accomplish this we train a control policy in an aircraft tracking task, explicitly including a measure of the estimator performance in the reward function. We also use a task-only policy trained with a reward function concerned only with the tracking task. Using simulations of agents following both policies we can evaluate the impact of rewarding estimator performance by comparing nominal task performance, estimator error, and measures of system observability.
\begin{figure}
    \centering
    \includegraphics{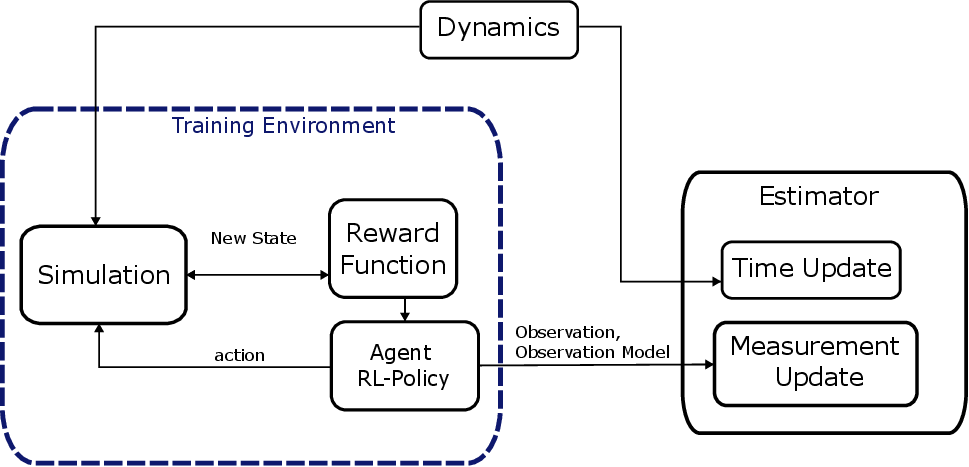}
    \caption{A control policy used to generate steering actions is trained using reinforcement learning. As the control is used to estimate the agent's state, this policy also forms the estimator's measurement function.}
    \label{fig:dynamics_rl_estimator_diagram}
\end{figure}
Figure \ref{fig:dynamics_rl_estimator_diagram} summarizes the general connections between the agent training, the system dynamics, and the estimator. In both training and evaluation, the estimator is implemented using the agent's system dynamics, the sampled pseudocontrol actions $\tilde{\mathbf{u}}$ as observations, and the learned policy $\pi$, as an observation model.

\subsection{Aircraft Goal Point Tracking}

We adopt a simple problem where a fixed-wing UAS must stay near a target on the ground. This task is representative of missions including persistent surveillance, environmental or scientific monitoring, and communication relay. Since the aircraft cannot hover, it must continuously maneuver to remain close to the target.
Assuming level flight and neglecting the rotational dynamics, the aircraft can be modeled as a unicycle kinematic model with states:
\begin{equation*}
    \mathbf{x} = 
    \begin{bmatrix}
        x_{aircraft} & y_{aircraft} & V & \psi
    \end{bmatrix}^\intercal
\end{equation*}
where $x_{aircraft}$ and $y_{aircraft}$ are the aircraft planar position states, $V$ is the aircraft inertial speed, and $\psi$ is the aircraft heading angle. The system dynamic equations for simulating the aircraft are given by: 

\begin{equation*}
    \frac{\partial \mathbf{x}}{\partial t} =
    \begin{bmatrix}
        V\cos(\psi)\\
        V\sin(\psi)\\
        u_{a} \\
        \frac{g \tan(u_{\phi})}{V}
    \end{bmatrix}
\end{equation*}
where $u_a$ is the speed rate of change command (set to zero for a constant-speed aircraft), $u_{\phi}$ is the bank angle command, and $g$ is the acceleration due to gravity. We assume steady turns and that the aircraft roll dynamics are fast, allowing the unicycle turn rate to be expressed as a bank angle which is more natural for fixed-wing aircraft (and which is more readily observed externally).

The decision state of the agent is the relative aircraft position and velocity to the target. For this study, the target is placed at the center of the environment (0,0) with zero velocity and the agent is set to a constant velocity. The agent state is thus directly related to the aircraft state by: 

\begin{equation}\label{eq:agent_state}
    \mathbf{s} = 
    \begin{bmatrix}
        x \\
        y \\
        v_x \\
        v_y
    \end{bmatrix} =
    \begin{bmatrix}
        x_{aircraft} -x_{target} \\
        y_{aircraft} -y_{target}\\
        V\cos(\psi) -v_{x_{target}}\\
        V\sin(\psi) -v_{y_{target}}
    \end{bmatrix}=
    \begin{bmatrix}
        x_{aircraft} \\
        y_{aircraft} \\
        V\cos(\psi) \\
        V\sin(\psi) 
    \end{bmatrix}
\end{equation}

where $x$ and $y$ are the position components relative to the target, $v_x$ and $v_y$ are the aircraft relative velocity components to the target, and $\psi$ is the aircraft heading angle.
The control policy takes the agent state as input and outputs a corresponding bank angle $u_{\phi}$ to control the aircraft. While it is possible to use the same state for simulation and the decision agent, it can be advantageous to distinguish between them. Frequently, the state-space required for simulation and that used for control differ. Here, the unicycle model captures the non-holonomic nature of the fixed-wing directional dynamics for simulation. However, for target tracking the relative position and velocity of the aircraft to the target is a more straightforward state definition. Defining the agent state in this way also avoids discontinuities in the agent's state-space when heading angles are near $-\pi$ or $\pi$. The utility of distinguishing between the simulation and agent state becomes especially clear when considering that this policy could be used to provide navigation guidance to a six degree of freedom simulation. 


\subsection{Agent Formulation and Training}

The agent's network architecture consists of an input layer, one 256-neuron hidden layer, and an output layer. The dimension of the input is equal to that of the agent's state. The pseudocontrol distribution is a Gaussian whose mean $\mu_{\mathrm{pseudo}}$, is obtained from the output of the policy network. 
In principle the variance could also be predicted by the network, but in practice this often resulted in very noisy control and poor estimator performance. To avoid this noisy control, while ensuring adequate exploration of the policy space, we set $\sigma$ of this distribution to $\sigma_{\mathrm{pseudo}}=\mathrm{arctanh}(\frac{5^\circ}{30^\circ})$ where 30$^\circ$ is the maximum allowed bank angle and 5$^\circ$ is the specified control action standard deviation related through Equation \ref{eq:pseudo_mapping}. The pseudocontrol distribution is then sampled, passed through a hyperbolic tangent function that maps its value to a range (-1,1), the output of which is scaled by the control limit, and used to step the agent dynamics.

The control objective in this formulation must be introduced through the reward function, which is designed to be large near the target, and decaying with range. 
We implement two reward functions, a task-only reward that aims to keep the aircraft near the target and an augmented reward that adds a term for estimator performance as outlined in Equation \ref{eq:embedded_reward}. 

\subsubsection{Task-only Reward}
We train a task-only policy using a reward which decays exponentially with distance between the aircraft position and the target.
\begin{equation}
    r_{\rm{task-only}} = \exp\left(-\frac{d^2 }{R_0^2}\right)\Delta t
    \label{eq:task-only_reward}
\end{equation}
where $R_0$ is a distance decay scalar factor, $d$ is the distance from the agent to the target ($d = (x^2 + y^2)^{\frac{1}{2}}$), and $\Delta t$ is the simulation step time size. Maximum possible reward is achieved when the aircraft is directly above the target location (0,0).

A potentially high-scoring control policy is one which maintains constant reward by flying a minimum radius circular path near the target. A simple thought experiment suggests that an estimator would have difficulty estimating the true state under this policy. Consider the aircraft flying in a circle centered on the target. At every point on this circle, the bank angle is identical. Observability requires that a change in state will result in a different sequence of observations, if the aircraft were displaced to another point on the circle then the resulting sequence of (constant) bank angles would be unchanged. We would then expect the estimator to determine that the aircraft lies on this circle, but not to accurately estimate the exact relative position and velocity of the aircraft to the target.

This also suggests that under such a policy that the initial sequence of commands which steer the aircraft to the circle may be more informative than those which maintain the circular path. These more informative observations allow the estimator to track the aircraft's relative position and velocity to the target, but once the commands become less informative at the circular path, the estimator struggles to distinguish the different states that produce the same bank angle commands.


\subsubsection{Embedded Estimator}
To produce policies which are more easily estimated, we also train a policy using an augmented reward function that includes the estimator's performance. This reward function takes the form shown in Equation \ref{eq:explicit_reward}:  

\begin{equation}
    r_{\mathrm{embedded}} = r_{\mathrm{task-only}} + \exp \left( -\frac{e^\intercal \Sigma_e e }{R_1^2}\right) \Delta t
    \label{eq:embedded_reward}
\end{equation}
where $R_1$ is a decay scale term, $e$ is the difference between true and the estimated state $e = \mathbf{s} \ - \hat{\mathbf{s}}$, and $\Sigma_e$ is a matrix weighting the error of each state. This reward function maximizes when the norm of the error between the true and estimated states is 0. 

\subsubsection{Training Environment}
The policy training is executed in OpenAi Gym \cite{brockman2016openai}, which allows developing reinforcement learning algorithms in a simulation environment. The neural network policy is implemented using the PyTorch package \cite{NEURIPS2019_9015} and optimized using the soft actor critic algorithm \cite{Haarnoja2019,SAC} for continuous action spaces. The training environment is a custom environment restricted to a two-dimensional plane, with dimensions of 500 m by 500 m. The action and observation spaces are summarized in Table \ref{tab:action_space} and \ref{tab:observation_space} respectively. Note that the observation space is defined larger than the simulation environment to ensure that the policy is valid even if the estimated position briefly exits the test environment.

In each training episode the aircraft is randomly placed in the environment at least one turn radius from the boundary. 
The aircraft speed is sampled uniformly between 10 $\mathrm{m\ s^{-1}}$ and 20 $\mathrm{m\ s^{-1}}$ and remains constant for each episode. The heading angle is sampled uniformly between $-\pi \mathrm{\ rad}$ and $\pi \mathrm{\ rad}$. The initial state boundaries for each training episode are summarized by the observation space in Table \ref{tab:ini_training_cond}.

\begin{table}[h]
    \centering
    \caption{Action space.}
    \label{tab:action_space}
    \begin{tabular}{ccc}
        \hline\hline
        Action & min & max \\
        \midrule 
        $u_{\phi} (\mathrm{rad})$ & $- \frac{\pi}{6}$ & $\frac{\pi}{6}$ \\
        \hline\hline
    \end{tabular}
\end{table}

\begin{table}[h]
    \centering
    \caption{Observation space.}
    \label{tab:observation_space}
    \begin{tabular}{ccc}
        \hline\hline
        Observation & min & max \\
        \midrule 
         $x (\mathrm{m})$ & $-500$ & $500$ \\
         $y (\mathrm{m})$ & $-500$ & $500$ \\
         $v_x (\mathrm{m\ s}^{-1})$ & $-25$ & $25$ \\
         $v_y (\mathrm{m\ s}^{-1})$ & $-25$ & $25$ \\
        \hline\hline
    \end{tabular}
\end{table}
\begin{table}[h]
    \centering
    \caption{Initial state conditions.}
    \label{tab:ini_training_cond}
    \begin{tabular}{ccc}
        \hline\hline
        Observation & min & max \\
        \midrule 
         $x (\mathrm{m})$ & $-205$ & $205$ \\
         $y (\mathrm{m})$ & $-205$ & $205$ \\
         $V (\mathrm{m\ s}^{-1})$ & $10$ & $20$ \\
         $\psi (\mathrm{rad})$ & $- \pi$ & $\pi$ \\
        \hline\hline
    \end{tabular}
\end{table}

Episodes have a time step size of 0.5 seconds and continue for 200 seconds or until the aircraft exits the environment.
No terminal reward is provided. We train the task-only and embedded policies using the same environment, true agent initialization boundaries, speed, maximum bank angle, episode duration, and step size.

\subsubsection{Selecting Reward Scale Factors}

The scale factors $R_0$ and $R_1$ in Equations \ref{eq:task-only_reward} and \ref{eq:embedded_reward} control how rapidly the reward function decays as the tracking or estimator error increases. In the tracking task, scale factor $R_0$ tunes the convergence of the aircraft to the vicinity of the target. When embedding the policy in the training signal, the scale factors are used to balance reward between the tracking task and estimator performance.

In previous work we observed that under the task-only reward (Equation \ref{eq:task-only_reward}), the policy tends to maximize its reward by flying directly to the target and engaging in a tight orbit centered on the target \cite{enriquez2024enabling}. This suggests that $R_0$ should be set based on the distance to the target that satisfies the tracking requirements, so that closer maneuvering is possible, but receives little additional reward. To avoid control saturation, a nominal task reward is established at a distance to target equal to the turn radius achieved at half the maximum bank angle. The initial value for $R_0$ is set so that at this distance from the target, the task reward is equal to $1-\exp(-1)$.

Similarly, $R_1$ is chosen to specify the reward at a nominal value for estimator error. Based on prior experiments using an ad hoc estimator error reward \cite{enriquez2024enabling} we chose an initial value of 15 with $\Sigma_e$ set to the identity matrix so that position and velocity states are equally weighted (because the velocity and position states both appear in the error there is not a dimensional interpretation to the error). As with $R_0$, the initial value for $R_1$ is obtained by setting the reward to $1-\exp(-1)$ at the nominal error.

While some further tuning of the scale factors was required, this structure allowed an informed choice of initial scale factors. Adjustments were then made in training by computing the relative task and estimator rewards, then adjusting the gains so that the rewards were approximately balanced with neither saturated. Final values for the scale factors are given in Table \ref{tab:scale_factors}.

\begin{table}[h]
    \centering
    \caption{Scale Factors.}
    \label{tab:scale_factors}
    \begin{tabular}{cc}
        \hline\hline
        Scale factor & value \\
        \midrule 
         $R_0$ & $121.8$ \\
         $R_1$ & $31.59$ \\
         \hline\hline
    \end{tabular}
\end{table}

\subsubsection{Policy Training}
Computing the estimator reward requires including the estimator in the training environment. By embedding the estimator in the agent model and including its performance in the reward signal we are directly optimizing over the estimator performance and, by extension, implicitly optimizing over observability of the resulting policy.

\begin{figure}
    \centering
    \includegraphics{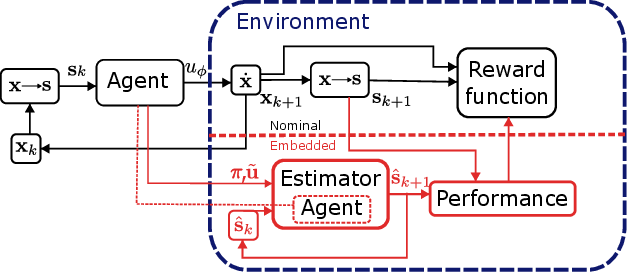}
    \caption{Training diagram for task-only and embedded policies.}
    \label{fig:training_diagram}
\end{figure}

Figure \ref{fig:training_diagram} summarizes the training process for the task-only and embedded estimator policies. For the task-only case (black boxes), an initial state $\mathbf{x}_k$ is sampled from initial aircraft state distribution per Table \ref{tab:ini_training_cond}. At each time step the agent's state is computed from the aircraft state (Equation \ref{eq:agent_state}). A bank angle $u_{\phi}$ is obtained from the agent after sampling a pseudocontrol $u$ from the policy's output distribution, passing it though the saturation function, and scaling it by the control angle limit per Equation \ref{eq:pseudo_mapping}. The bank angle $u_{\phi}$ is used to simulate the aircraft state forward with the aircraft dynamics. The state is converted back to the next agent state $\mathbf{s}_{k+1}$ to calculate the task reward. For the task-only case, maximizing this reward function is the agent's training objective. The embedded estimator policy adds the red boxes to the training process. The estimator takes the agent, the sampled pseudocontrol $\tilde{u}$, and the initial estimated state $\hat{\mathbf{s}}_k$ given by the initial aircraft relative position and velocity to the target, to produce the next estimated agent state $\hat{\mathbf{s}}_{k+1}$. The estimated and true agent state $\mathbf{s}_{k+1}$ are used to compute the estimator error performance $e$. The reward function becomes the embedded reward in Equation \ref{eq:embedded_reward}

During training, the estimator state is initialized to the true agent state plus a randomly selected distance and velocity error, sampled from a Gaussian distribution. This initialization error is centered at the true state with a standard deviation of 50 m for each position component and 5 $\mathrm{m\ s^{-1}}$ for each velocity component.

To determine when a policy has converged, the reward accumulated in each episode is compared to a threshold performance. For the task-only policy this threshold is defined:
\begin{equation*}
    R_{threshold,task} = (0.85)\exp\left(-\frac{\mathrm{minimum aircraft turn radius}^2}{R_0^2}\right) t
\end{equation*}
where the minimum aircraft turn radius $\frac{V^2}{g\tan u_{\phi \mathrm{max}}}=$ 45.2 m at maximum bank angle command and nominal speed and $t$ is the episode duration time. For the policy which embeds the estimator, the threshold reward is set to:
\begin{equation*}
R_{threshold,embedded} = (0.85)\left(\exp\left(-\frac{\mathrm{minimum aircraft turn radius}^2}{R_0^2}\right) +\exp\left(-\frac{\mathrm{expected error performance}^2}{R_1^2}\right)\right) t
\end{equation*}
with expected error performance set to 15 as described for the $R_1$ scale factor. The factor of 0.85 is applied as both the task and estimator performance may take some time after initialization to converge, so achieving 100\% of the nominal reward is not expected.

The training for each policy is terminated when the average of the accumulated reward in the last 100 episodes reaches the threshold.
Applying the corresponding threshold resulted in 2,507 training episodes (493,127 simulation steps) for the task-only reward case, and 6,115 training episodes (2,354,219 simulation steps) for the embedded estimator reward case. 
The embedded estimator reward case is expected to have more training as it is a more complicated reward.

%% file: results_v2_v2.tex
\section{Results}

The task-only and embedded estimator reward cases are evaluated using an ensemble of 14,500 episodes, generated for a fixed policy after completion of training. To initialize each episode, the aircraft state is sampled uniformly, excluding position states within one turn radius of the boundary. In all cases, the aircraft speed is set to $16~\mathrm{m~s^{-1}}$. The initialization range for the true state initial is summarized in Table \ref{tab:ini_training_cond}, excluding the initial speed $V$, which is set to $16~\mathrm{m~s^{-1}}$ for all episodes. The aircraft state is converted to the agent state as per Equation \ref{eq:agent_state}. The estimator is initialized at the true agent state, plus a distance and velocity error sampled from a Gaussian distribution, centered at the true state with a standard deviation of 50 m for each position component, and $5~ \mathrm{m~s^{-1}}$ for each velocity component. The estimator covariance is initialized consistent with this error distribution. The $\sigma$ values for the Gaussian distributions used for sampling the initial condition of the estimated state, are summarized in Table \ref{tab:ini_estimated}, where all distributions are centered at the true state.

\begin{table}[h]
    \centering
    \caption{Estimated state $\sigma$ for initial condition distribution sampling.}
    \label{tab:ini_estimated}
    \begin{tabular}{cc}
        \hline\hline
        Estimated state & $\sigma$ \\
        \midrule 
         $x\ (\mathrm{m})$ & $50$  \\
         $y\ (\mathrm{m})$ & $50$  \\
         $v_x\ (\mathrm{m\ s}^{-1})$ & $5$  \\
         $v_y\ (\mathrm{m\ s}^{-1})$ & $5$ \\
        \hline\hline
    \end{tabular}
\end{table}

\begin{figure}[h]
    \centering
    \begin{minipage}[b]{.49\textwidth}
    \includegraphics[width=\textwidth]{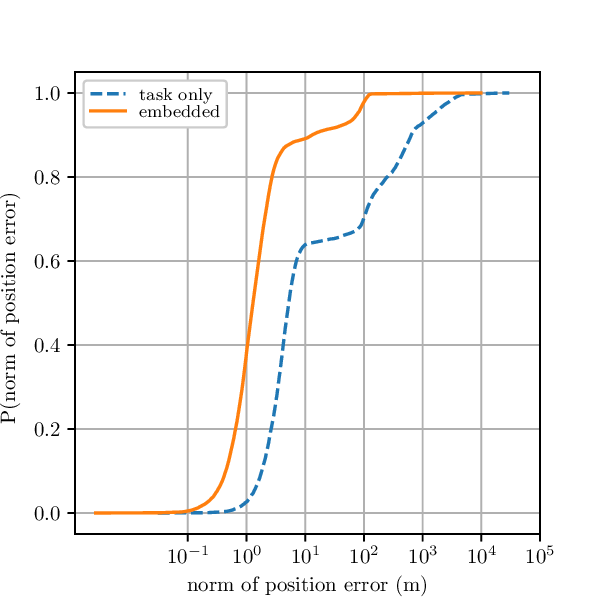}
    \caption{ECDF of the log of the norm of the position error at the simulation end time.}
    \label{fig:CDF_pos_error_norm_full}
    \end{minipage}
    \hfill
    \centering
    \begin{minipage}[b]{.49\textwidth}
        \includegraphics[width=\textwidth]{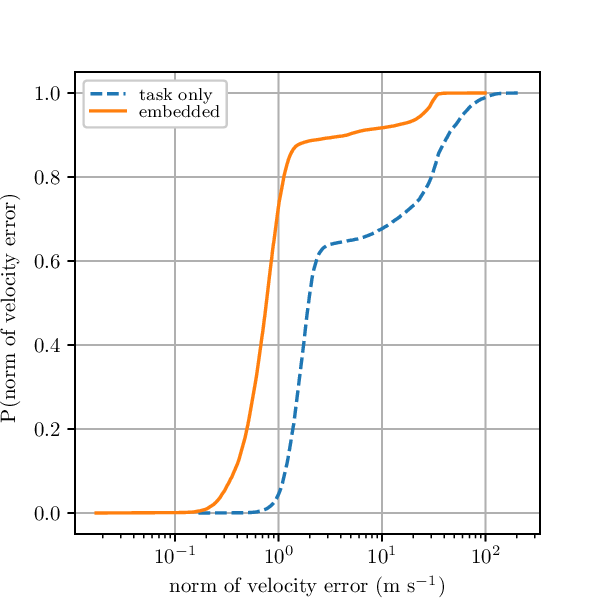}

    \caption{ECDF of the log of the norm of the velocity error at the simulation end time.} 
    \label{fig:CDF_vel_error_norm_full}
    \end{minipage}
\end{figure}

The resulting empirical cumulative distributions (ECDF) for the norm of the position and velocity error at 200 seconds of simulation time (end of the episode), are shown in Figures \ref{fig:CDF_pos_error_norm_full} and \ref{fig:CDF_vel_error_norm_full} respectively. 
Figure \ref{fig:CDF_pos_error_norm_full} shows that using the augmented control policy reduces error by a factor of approximately four at the median performance level.

When considering the tail of poor performance, the augmented policy achieves less than 10 m of error in about 90 \% of the cases while task-only policy can only achieve this level of error in 60 \% of cases.
Figure \ref{fig:CDF_vel_error_norm_full} displays the norm of the velocity error, showing that the augmented policy is able to reduce the velocity error relative to the task-only policy at every percentile point with a pattern similar to the position error.

\begin{figure}[h]
    \centering
    \begin{minipage}[b]{.49\textwidth}
    \includegraphics[width=\textwidth]{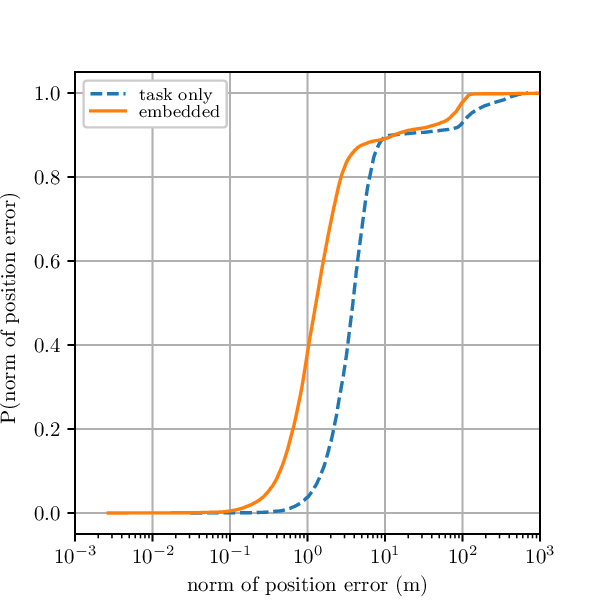}
    \caption{ECDF of the log of the norm of the position error at simulation end time when task-only cases which diverge are removed.}
    \label{fig:CDF_pos_error_norm}
    \end{minipage}
    \hfill
    \centering
    \begin{minipage}[b]{.49\textwidth}
    \includegraphics[width=\textwidth]{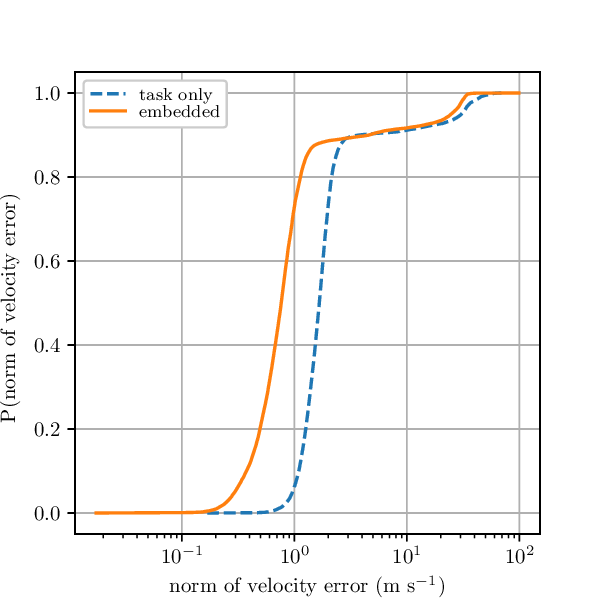}
    \caption{ECDF of the log of the norm of the velocity error at simulation end time when task-only cases which diverge are removed.}
    \label{fig:CDF_vel_error_norm}
    \end{minipage}
\end{figure}

The difference in estimator performance under these two policies is striking, but driven in part by divergence of a significant fraction of the task-only policy simulations. 
This in itself is significant, but a best-case comparison can be constructed by excluding runs where the estimator diverged.
To evaluate this, episodes where the estimated state exits a 1 km region centered on the objective are excluded (twice the environmental boundary) for the task-only case only (very few of the embedded policy informed cases exceed this threshold). 
This results in 4,618 episodes removed (31.8\% of the total number of episodes). Figures \ref{fig:CDF_pos_error_norm} and \ref{fig:CDF_vel_error_norm} show the ECDF for the task-only policy evaluations subset against the full set of augmented policy evaluations.

Figure \ref{fig:CDF_pos_error_norm} shows that the embedded case outperforms the task-only case at almost all percentiles with a very small crossover near 10 m of error.
At the median level, the norm of the position when using the task-only policy is 3.85 m, while the position error when using the embedded estimator policy is 1.29 m, a 66.5\% improvement over the task-only policy. 

At error percentiles of 90\% or greater the two estimators perform similarly.
Figure \ref{fig:CDF_vel_error_norm} shows the ECDFs for the norm of the velocity error which follows a trend very similar to the position performance.

\begin{figure}[h]
    \centering
    \begin{minipage}[b]{.49\textwidth}
        \includegraphics[width=\textwidth]{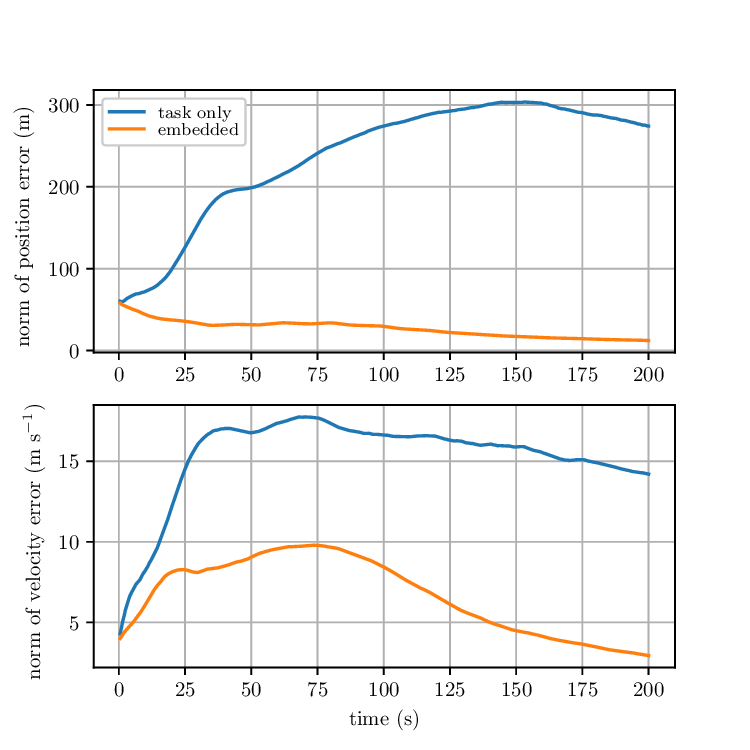}
    \caption{Time history of the ensemble mean of the norm of the estimator error including all episodes}
    \label{fig:mean_error_norm_full}
    \end{minipage}
    \hfill
    \centering
    \begin{minipage}[b]{.49\textwidth}

    \includegraphics[width=\textwidth]{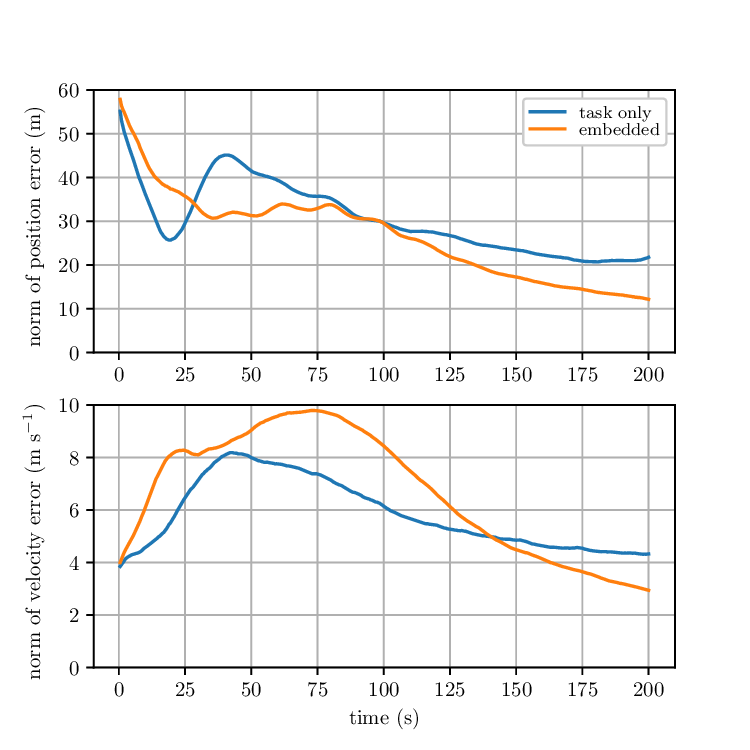}
    \caption{Time history of the ensemble mean of the norm of the estimator error excluding episodes which diverge.}
    \label{fig:mean_error_norm}
    \end{minipage}
\end{figure}

Figures \ref{fig:mean_error_norm_full} and \ref{fig:mean_error_norm} show the time history of the norm of the position and velocity errors averaged over all simulation episodes.
In the task-only case the position error degrades immediately, recovering only slightly after 160 seconds of simulation. In contrast, the mean estimator error in the embedded case decreases continuously until the end of the simulation. At 200 seconds, the mean of the norm of the position error is 274.3 m for the task-only reward case and 12.15 m for the embedded estimator reward case. This difference represents a 95.6\% decrease in error for the embedded estimator reward case. The velocity mean error norm increases for both the task-only and embedded cases and start to decrease after 75 seconds. The mean of the norm of the velocity error at 200 seconds is 14.2 m s$^{-1}$ for the task-only reward case and 2.94 m s$^{-1}$ for the embedded estimator case. This difference represents a 79.3 \% improvement in estimator performance.

The mean error is dominated by the 30\% of task-only cases in which the estimator diverges. Figure \ref{fig:mean_error_norm} shows the time history of the mean error, excluding episodes where the estimated state position exceeds the simulation region boundaries. 
The task-only and embedded estimator reward cases have similar convergence behavior for the initial 15 seconds. This is approximately the time required for the aircraft to reach the target location. 
After 15 seconds, the task-only case mean error increases to 45 m, then slowly decreases for almost the remainder of the simulation. 
The embedded case mean error decreases to approximately 30 m, holding steady until around 80 seconds, then decreases until the end of the simulation. 
At 200 seconds, the mean of the norm of the position error is 21.77 m for the task-only reward case and 12.15 m for the embedded estimator reward case. This difference represents a 44.2\% decrease in error for the embedded estimator reward case. The mean of the norm of the velocity error at 200 seconds is 4.33 $\mathrm{m~s^{-1}}$ for the task-only reward case and 2.94 $\mathrm{m~s^{-1}}$ for the embedded estimator reward case. This reduction represents a 32.1\% decrease in error for the embedded estimator reward case. 

\begin{figure}[h]
    \centering
        \includegraphics[scale=1.0]{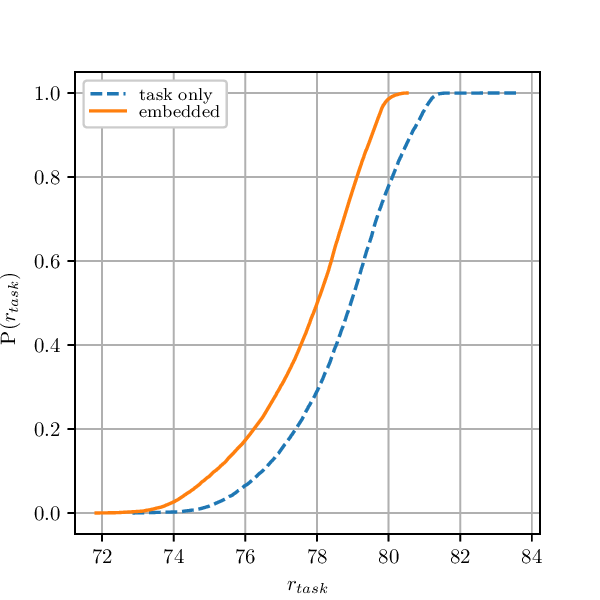}

    \caption{ECDF of the task reward obtained per episode for the task-only and embedded estimator policies. The task-only reward case performs slightly better in the task across all percentile points.}
    \label{fig:distance_reward}
\end{figure}

Given that the augmented reward function allows a reward greater than the task-only case even if the task performance is worse (because of the estimator ``bonus''), a potential concern is that the policy trained to minimize estimator error trades this for a reduction in task performance.
The tracking reward ($r_{task-only}$ in Equations \ref{eq:task-only_reward} and \ref{eq:embedded_reward}) measures performance on the mission objective of remaining near the fixed target location. It is desirable that producing a more observable policy does not compromise the tracking task performance. To evaluate this effect, Figure \ref{fig:distance_reward} shows the cumulative distribution of tracking reward per episode. For the full evaluations, both the task-only and the embedded policies show close performance across all percentile points. At the 80th percentile, the rewards are 80.1 for the task-only case and 79.1 for the embedded case, a difference of 1.2\%.

\begin{figure}[h]
    \centering
        \includegraphics[width=\textwidth]{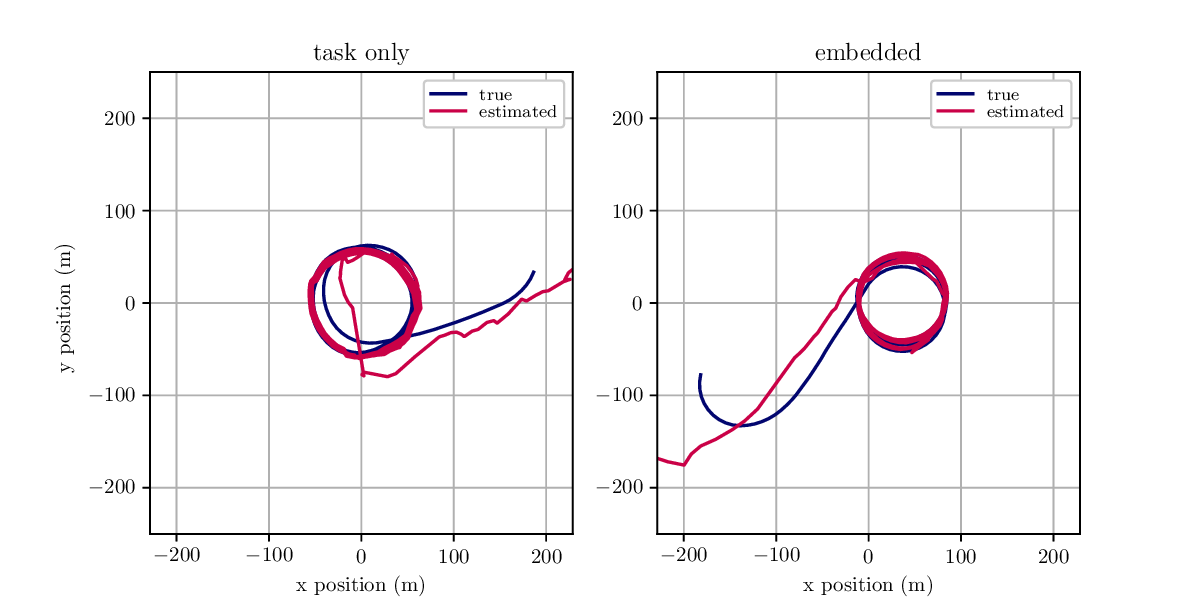}
    \caption{True and estimated state trajectories for a single run. Left: task-only reward case. Right: embedded estimator reward case.}
    \label{fig:trajectories}
\end{figure}

Figure \ref{fig:trajectories} illustrates true and estimated state trajectories from a single run for both reward cases. Both reward cases are initialized at the conditions corresponding to median performance from the ensemble of 14,500 simulations. The task-only and embedded reward cases both fly roughly a circular trajectory. This circle is centered roughly at the origin for the task-only case, but displaced slightly to the east in the embedded case.

\begin{figure}[h]
    \centering
        \includegraphics[width=\textwidth]{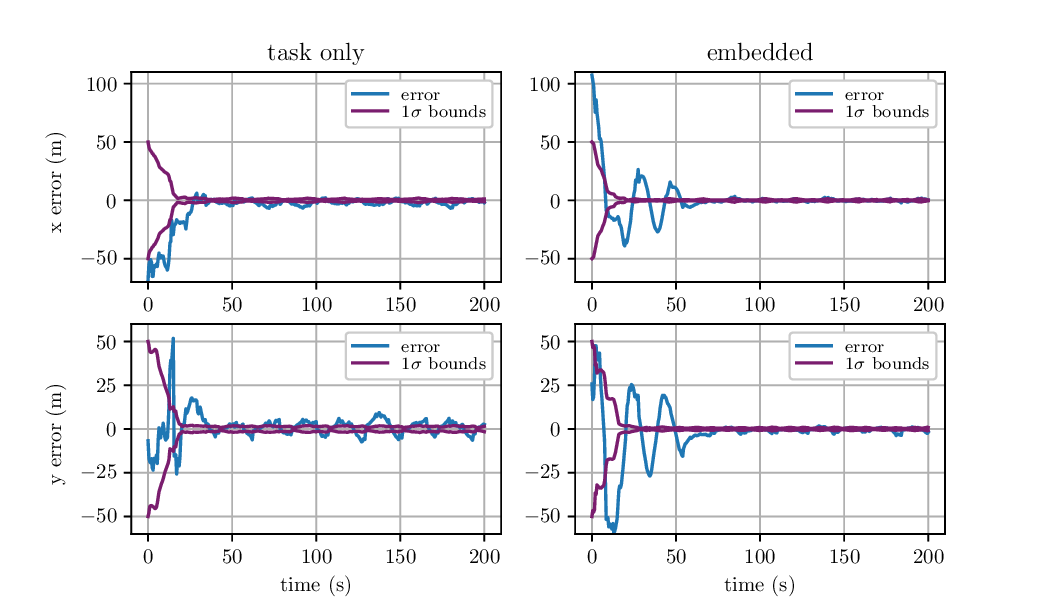}
    \caption{Time history and $1\sigma$ bounds of the position error for the task-only and embedded estimator reward cases from an example run.}
    \label{fig:pos_error}
\end{figure}

The corresponding time history of the position error is shown in Figure \ref{fig:pos_error}. The initial x-position error is 68.3 m for the task-only case, and 107.2 m for the embedded case. The initial y-position errors are 6.7 m and 25.8 m for the task-only and embedded cases, respectively. The $1\sigma$ bounds for the x-position error of the task-only reward case settle at 0.5-2.0 m. For the embedded estimator reward case, the $1\sigma$ bounds range reduces to 0.2 - 1.4 m. The $1\sigma$ bounds for the y-position error of the task-only reward case settle at 0.8-2.0 m. For the embedded estimator reward case, the $1\sigma$ bounds range reduces to 0.3 - 1.0 m. For the task-only reward case, the error is within the $1\sigma$ bounds 20.7\%, and 25.9\% of the time for the x and y position errors, respectively. For the embedded estimator reward case, the error is within the $1\sigma$ bounds 33.2\%, and 31.9\% of the time for the x and y position errors, respectively.

Since the initial position error for each case is different, it is more informative to compare the mean error for the last 100 seconds of simulation. For the task-only case, the x and y position mean error for the last 100 seconds is 2.24 m and 2.62 m respectively. For the embedded case, the x and y position mean error for the last 100 seconds is 0.64 m and 0.85 m respectively. Table \ref{tab:position_error} summarizes the position error results.

\begin{table}[h]
    \centering
    \caption{Position state error}
    \label{tab:position_error}
    \begin{tabular}{cccc}
        \hline\hline
        Case & mean error(m) (last 100 sec.) & $1 \sigma$ bounds(m) & $1 \sigma$ bounds inside(\%) \\
        \midrule 
        $x-$task-only & 2.24 & 0.5-2.0 & 20.7\\
        $x-$embedded & 0.64 & 0.2-1.4 & 33.2 \\
        $y-$task-only & 2.62 & 0.8-2.0 & 25.9\\
        $y-$embedded & 0.85 & 0.3-1.0 & 31.9\\
        \hline\hline
    \end{tabular}
\end{table}

\begin{figure}[h]
    \centering
        \includegraphics[width=\textwidth]{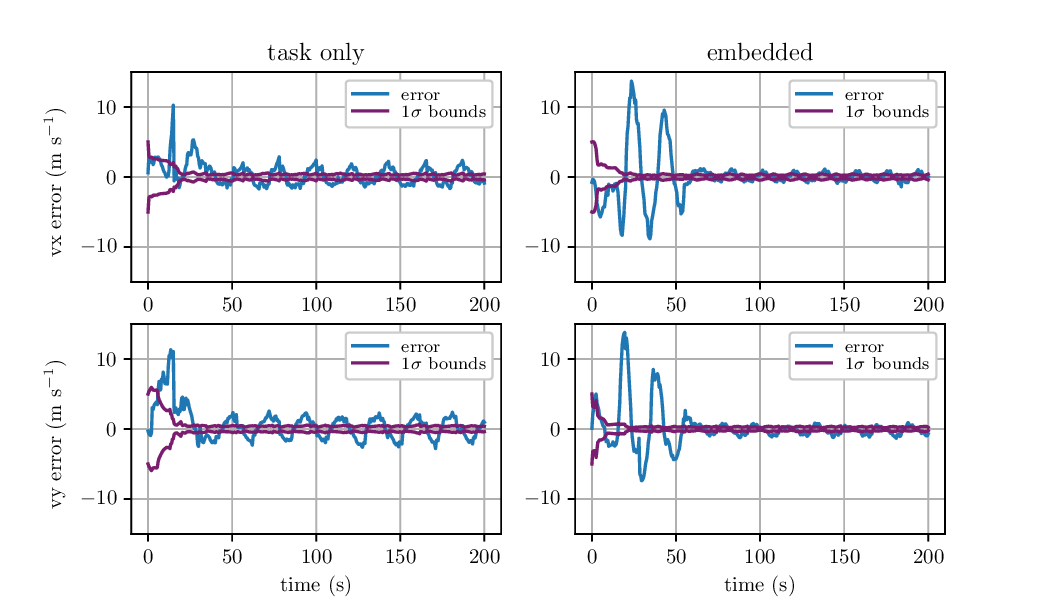}
    \caption{Time history and $1\sigma$ bounds of the velocity error for the task-only and embedded estimator reward cases from an example run.}
    \label{fig:vel_error}
\end{figure}

The time history of the velocity error is shown in Figure \ref{fig:vel_error}. The initial x-velocity error for the task-only case is 0.62 m s$^{-1}$, and 0.76 m s$^{-1}$ for the embedded case. The initial y velocity error is 0.30 m s$^{-1}$ for the task-only case, and 0.09 m s$^{-1}$ for the embedded case. The $1\sigma$ bounds for the x-velocity error of the task-only reward case settle at 0.25-0.60 m s$^{-1}$. For the embedded estimator reward case, the $1\sigma$ bounds reduce to 0.20-0.45 m s$^{-1}$. The $1\sigma$ bounds for the y-velocity error of the task-only reward case settle at 0.25-0.55 m s$^{-1}$. For the embedded estimator reward case, the $1\sigma$ bounds reduce to 0.16-0.40 m s$^{-1}$. For the task-only reward case, the error is within the $1\sigma$ bounds 17.5\%, and 16.5\% of the time for the x and y velocity errors, respectively. For the embedded estimator reward case, the error is within the $1\sigma$ bounds 28.2\%, and 25.2\% of the time for the x and y velocity errors, respectively. 

For the task-only case, the x and y velocity mean error for the last 100 seconds is 1.01 m s$^{-1}$ and 1.11 m s$^{-1}$ respectively. For the embedded case, the x and y position mean error for the last 100 seconds is 0.43 m s$^{-1}$ and 0.45 m s$^{-1}$ respectively. Table \ref{tab:velocity_error} summarizes the velocity error results.

\begin{table}[h]
    \centering
    \caption{Velocity state error}
    \label{tab:velocity_error}
    \begin{tabular}{cccc}
        \hline\hline
        Case & mean error(m s$^{-1}$)(last 100 sec.) & $1 \sigma$ bounds(m s$^{-1}$) & $1 \sigma$ bounds inside(\%) \\
        \midrule 
        $x-$task-only & 1.01& 0.25-0.60 & 17.5\\
        $x-$embedded & 0.43 & 0.2-45 & 28.2 \\
        $y-$task-only & 1.11 & 0.25-0.55 & 16.5\\
        $y-$embedded & 0.45 & 0.16-0.40 & 25.2\\
        \hline\hline
    \end{tabular}
\end{table}

%% file: discussion_v2.tex
\section{Discussion}

The distribution of estimator performance for the embedded estimator reward case shows superior performance at all percentiles for the full evaluation set. Even when excluding evaluations in which the task-only case diverges, the estimator achieves better performance at almost all percentiles when the agent follows the policy rewarded for estimator performance. The frequency of filter divergence is significantly higher for the task-only case compared to the embedded estimator reward case. This filter divergence suggests that poor observability of the task-only policy allows the estimate to drift outside of the region in which the policy is trained, at which point the policy function is no longer valid as either a control or observation model.

The mean of the norm of the state error is significantly smaller for the estimator using the embedded estimator policy for both position and velocity for the entire simulation. When examining the mean error of only converged estimators (Figure \ref{fig:mean_error_norm}), the results show that the task-only policy can produce estimates which outperform those produced under the embedded policy early in the simulation, although as time goes on, the embedded policy produces significantly better estimates.

A transition in the performance of the estimators is observed around 20-30 seconds into the simulation -- approximately the time required for the aircraft to reach the target. It appears that the task-only case is fairly observable during the initial flight to the orbit, but becomes less so once the aircraft begins orbiting the target. The embedded case on the other hand achieves modest performance until the aircraft has completed several orbits.

In addition to reducing the error in the state estimate, the embedded policy achieves a more consistent estimate. Figures \ref{fig:pos_error} and \ref{fig:vel_error} indicate that the rate of estimates exceeding the $1-\sigma$ bounds for the embedded policy are more consistent with the state covariance estimate than for the task-only policy.

In prior results with an ad-hoc ``observable'' reward function in \cite{doi:10.2514/6.2023-2657}, there was a preferred trajectory for the embedded estimator reward case, which allowed the estimator to uniquely identify the true state.
In contrast, the task-only reward case showed roughly a circular path around the target. A simple thought experiment suggests that no state in the orbit can be distinguished based on bank angle in a constant bank angle orbit. These results, however, show nearly no clear distinction between the two reward cases other than the center of the orbit at different locations. The embedded policy achieves observability with only subtle changes to the shape of the actual trajectory. 

The improvement in state quality achieved by rewarding estimator performance suggests that the resulting policy must make the system more observable. 
Two notions of observability are relevant -- observability at each point, approximated by the linearized observability matrix (Equation \ref{eq:LOM}); and observability of the trajectory, approximated by the stripped observability matrix (Equation \ref{eq:SOM}). 
Table \ref{tab:singular_values} illustrates the observability at each point as diagnosed by taking the mean of the singular values of the linearized observability matrix over every point in the example case.
The smallest singular value is always practically zero for both task-only and embedded estimator reward cases. 
This means that at no point in the state-space visited by the aircraft is the system fully observable at that state.

\begin{table}[h]
    \centering
    \caption{Mean singular values of the linearized observability matrix for one test case.}
    \label{tab:singular_values}
    \begin{tabular}{ccc}
        \hline\hline
        Singular \\Value & task-only avg & embedded avg \\
        \midrule 
        1 & 0.744 & 39.873 \\
        2 & 0.072 & 1.012 \\
        3 & 0.007 & 0.031 \\
        4 & $3.973\times 10^{-10}$ & $2.769\times 101^{-8}$\\
        \hline\hline
    \end{tabular}

\end{table}

While neither policy is fully observable at any single point, the observability of the trajectories does differ. 

\begin{figure}[h]
    \centering
        \includegraphics[width=\textwidth]{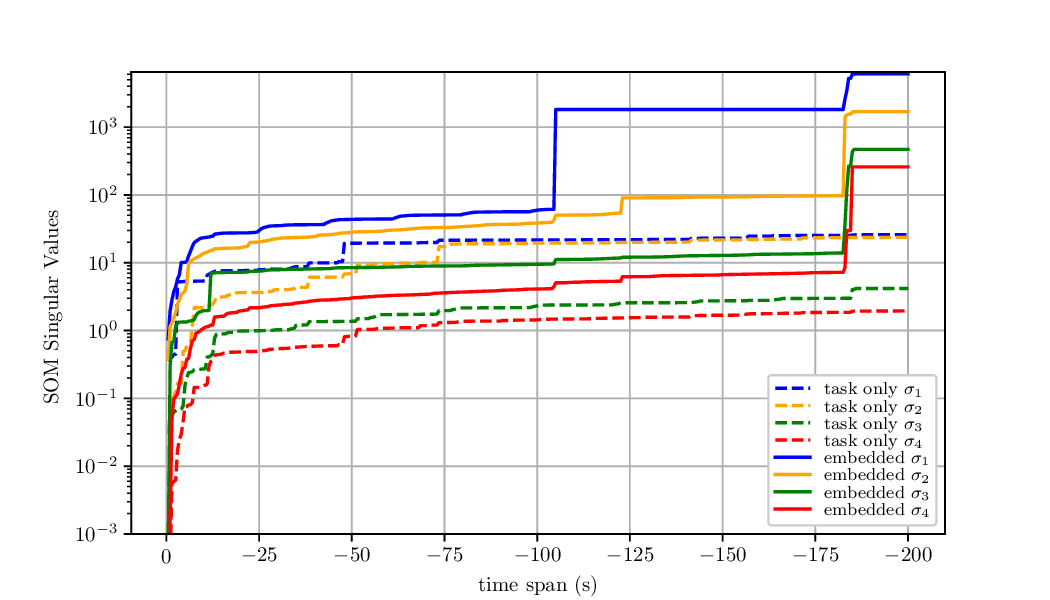}
    \caption{Singular values of SOM for increasing time spans starting at the end of the simulation for the task-only and embedded estimator reward cases. All singular values are larger for the embedded estimator reward case.  }
    \label{fig:timespan_singular_values}
\end{figure}

Figure \ref{fig:timespan_singular_values} compares the singular values of the stripped observability matrix for the two policies. The singular values are obtained from sections of the full SOM spanning increasing time spans in reversed order (end to beginning of simulation, so that a time span of -25 s includes the last 25 seconds of the simulation ). The time span increases by $0.5$ seconds at each iteration. The singular values are obtained in reverse order in order to better isolate the observability of the system for the aircraft behavior at the target. Once approximately 50 seconds (two cycles) of data are included in the SOM, the smallest singular value of the SOM for the embedded case increases from 0 to 1, indicating that the sequence is observable. While the full state is not observable at any given epoch, our interpretation is that the direction of the unobservable subspace is not constant, which allows a sequence of states to achieve observability. This analysis would indicate that a degree of observability is achieved for state sequences under both the task-only and embedded estimator reward cases, however, all singular values for the embedded estimator reward are larger, indicating a greater degree of observability. This aligns with the policy trained with the embedded estimator reward having superior estimator performance. These results confirm that by embedding the estimator in the training system, we can improve the observability of the system.

The task performance shows that there is a small ``penalty'' in augmenting the task-only reward with the estimator performance. While it is expected that the task-only reward case outperforms the embedded estimator reward case, previous experiments have shown reversed results \cite{doi:10.2514/6.2023-2657}. These task-only reward results suggest that there appear to be cases where a number of policies achieve nearly the same task performance. Inclusion of the estimator performance guides the training to one of these policies which also provides good estimator performance. If we conceptualize a policy-space, the two policies we examine lie nearly on the same locus of policies providing a given level of task reward, by introducing the estimator reward, we are able to select a policy which also provides observability. Drawing an analogy to a linear control context, this is akin to modifying a control vector with an auxiliary objective where the difference between nominal and augmented control lies in the null space of the control matrix.

Taken as a whole, the results suggest that integrating the estimator into the agent at the training stage and providing a reward signal for its performance provides a means to generate control policies whose input state can be determined from the output. The policy space appears to be large enough that the control policy can be made observable without significant impact on the task-specific performance. Embedding an estimator at training time thus appears to provide a path forward for enabling coordination between autonomous agents under limited communication or direct observation. While we have embedded an estimator at training time, we still achieve a degree of separation between agents allowing them to be trained independently, and to continue to coordinate under re-tasking provided the policy used by other agents is known. This stands in contrast to ``fully integrated'' approaches which train all agents simultaneously, implicitly developing the estimator as part of each agent's control policy such that it cannot be easily updated without retraining.

This work does not however, fully address the objective of enabling coordination between humans and autonomous systems. While human cognition exhibits Kalman-filter-like capabilities \cite{Wittmann2016,Miall2008,Blakemore2000}, human understanding of a system's state is not likely to be described by the UKF dynamics precisely. Regardless of the ``algorithm,'' observability can be viewed as an information-theoretic condition enabling reconstruction of state information from observations \cite{Mohler1988}. Thus, systems which are not observable cannot be made so regardless of the sophistication of the estimator. Even when a system is not precisely unobservable, such as the task-only reward policy here, an improvement in the observability should be expected to improve the quality of an estimate and to reduce the computational requirement (compute time or human cognitive effort) to obtain an estimate of given quality.

%% file: conclusion_v2.tex
\section{Conclusion}

Estimation of the state of decision-making agents given observations of their control decisions is difficult as the control policy is typically not a one-to-one map between states and controls and equilibrium controls often have relatively little observational diversity. To address this, we include an explicit measure of the estimator's performance in the reward function used to train a control policy using reinforcement learning. This policy's performance in estimating an agent's state when only the agent's actions are used as observations performed significantly better than a policy that does not include the estimator's performance in its reward function. This performance improvement suggests that coordination based on action observation alone might be possible, and that we can effectively \say{compress} the state information stream by using a policy that is intended to be observable. The policy space may be complex enough that observable policies achieve similar performance even with the addition of an observability objective.

Designing control policies for autonomous systems to be observable can make communication more efficient, or enable coordination when direct observation is possible but explicit communication is not.  In a human-machine teaming context, when control policies are not observable, we expect that humans will have difficulty understanding the state of autonomous systems (and perhaps anticipating their actions). Autonomous systems guided by observable policies may be a path to enabling coordination among large numbers of autonomous systems and between humans and autonomous systems.